\documentclass[11pt]{article}

\usepackage[final]{acl}

\usepackage{times}
\usepackage{latexsym}
\usepackage{multirow}

\usepackage[T1]{fontenc}
\usepackage[utf8]{inputenc}

\usepackage{microtype}

\usepackage{inconsolata}

\usepackage{graphicx}

\usepackage{hyperref}
\usepackage{xspace}
\usepackage{url}
\usepackage{booktabs}
\usepackage{multirow}
\usepackage[table]{xcolor}  % 支持 \cellcolor 和 HTML 颜色
\usepackage{booktabs}
\usepackage[table]{xcolor}
\usepackage{amsfonts}
\usepackage{array}      % p{..} + \raggedright
\usepackage{adjustbox}  % max width=\textwidth
\usepackage[table]{xcolor}
\usepackage[table]{xcolor}
\definecolor{matgreen}{RGB}{0,128,0}  % Matplotlib 的 'green' 端点
\usepackage{xcolor}
\usepackage{xparse}
\usepackage{graphicx}
\usepackage{wrapfig}
\usepackage[font=small,labelfont=bf]{caption} % 可选，美化caption

\definecolor{heatred1}{HTML}{FEE0D2} % very light
\definecolor{heatred2}{HTML}{FCAE92} % light (推荐用于正文高亮)
\definecolor{heatred3}{HTML}{EF4444} % medium
\definecolor{heatred4}{HTML}{B91C1C} % deep

\definecolor{matgreen}{HTML}{86EFAC} % 近似浅绿，可继续用你原本的

\NewDocumentCommand{\hlneg}{O{2}m}{%
  \begingroup\setlength{\fboxsep}{0.6pt}\colorbox{heatred#1}{\strut #2}\endgroup}
\NewDocumentCommand{\hlpos}{O{2}m}{%
  \begingroup\setlength{\fboxsep}{0.6pt}\colorbox{matgreen!35}{\strut #2}\endgroup}

\newcommand\ourmetric{\textsc{SustainScore}\xspace}
\usepackage{amsthm}
\theoremstyle{definition} % Or another style you prefer

\usepackage{graphicx}
\usepackage{subcaption}
\usepackage{tabularx}
\usepackage[fleqn]{amsmath}
\usepackage[most]{tcolorbox}
\usepackage{caption}
\usepackage{booktabs}
\usepackage{tabularx}
\usepackage{enumitem}

\title{On the Paradoxical Interference \\
between Instruction-Following and Task Solving}

\newcommand{\SP}{\textsuperscript{$\spadesuit$}}
\newcommand{\CL}{\textsuperscript{$\clubsuit$}}

\author{Yunjia Qi\SP, Hao Peng\SP, Xintong Shi\SP, Amy Xin\SP, \\ \textbf{Xiaozhi Wang\CL, Bin Xu\SP, Lei Hou\SP, Juanzi Li\SP} \\
        \SP Department of Computer Science and Technology, BNRist;\\
\CL Shenzhen International Graduate School;\\
Tsinghua University \\ \texttt{qyj23@mails.tsinghua.edu.cn}}

\begin{document}
\maketitle
\begin{abstract}
%%%%%%%%%%%%%%%

% Instruction following serves as a bridge between large language models' (LLMs) problem-solving capabilities and human intent, enabling users to specify not just the goal of a task, but the specific means of achieving it. 
% However, we find that restricting how a task is solved often hurts what is achieved, similar to a phenomenon known as goal displacement effect in sociology.

Instruction following aims to align Large Language Models (LLMs) with human intent by specifying explicit constraints on how tasks should be performed. 
However, we reveal a counterintuitive phenomenon: instruction following can paradoxically interfere with LLMs' task-solving capability. We propose a metric, \textit{\ourmetric}, to quantify the interference of instruction following with task solving. 
It measures task performance drop after inserting into the instruction a self-evident constraint, which is naturally met by the original successful model output and extracted from it. 
% It measures how much task performance is sustained when a self-evident constraint, naturally met by the original model output, is inserted into the instruction.
% It measures the task performance drop after inserting a self-evident constraint into the instruction, which is naturally met by the original model output and extracted from it. 
Experiments on current LLMs in 
mathematics, multi-hop QA, and code generation
show that adding the self-evident constraints leads to substantial performance drops, even for advanced models such as Claude-Sonnet-4.5.
We validate the generality of the interference across constraint types and scales.
Furthermore, we identify common failure patterns, and by investigating the mechanisms of interference, we observe that failed cases allocate significantly more attention to constraints compared to successful ones.
Finally, we use \ourmetric to conduct an initial investigation into how distinct post-training paradigms affect the interference, presenting empirical observations on current alignment strategies.
We will release our code and data to facilitate further research\footnote{https://github.com/kijlk/IF-Interference}.

\end{abstract}

\section{Introduction}

% alignment是很重要的，现在有很多alignment的方式来完成
\begin{figure}[t]
    \centering
    \includegraphics[width=1.0\linewidth]{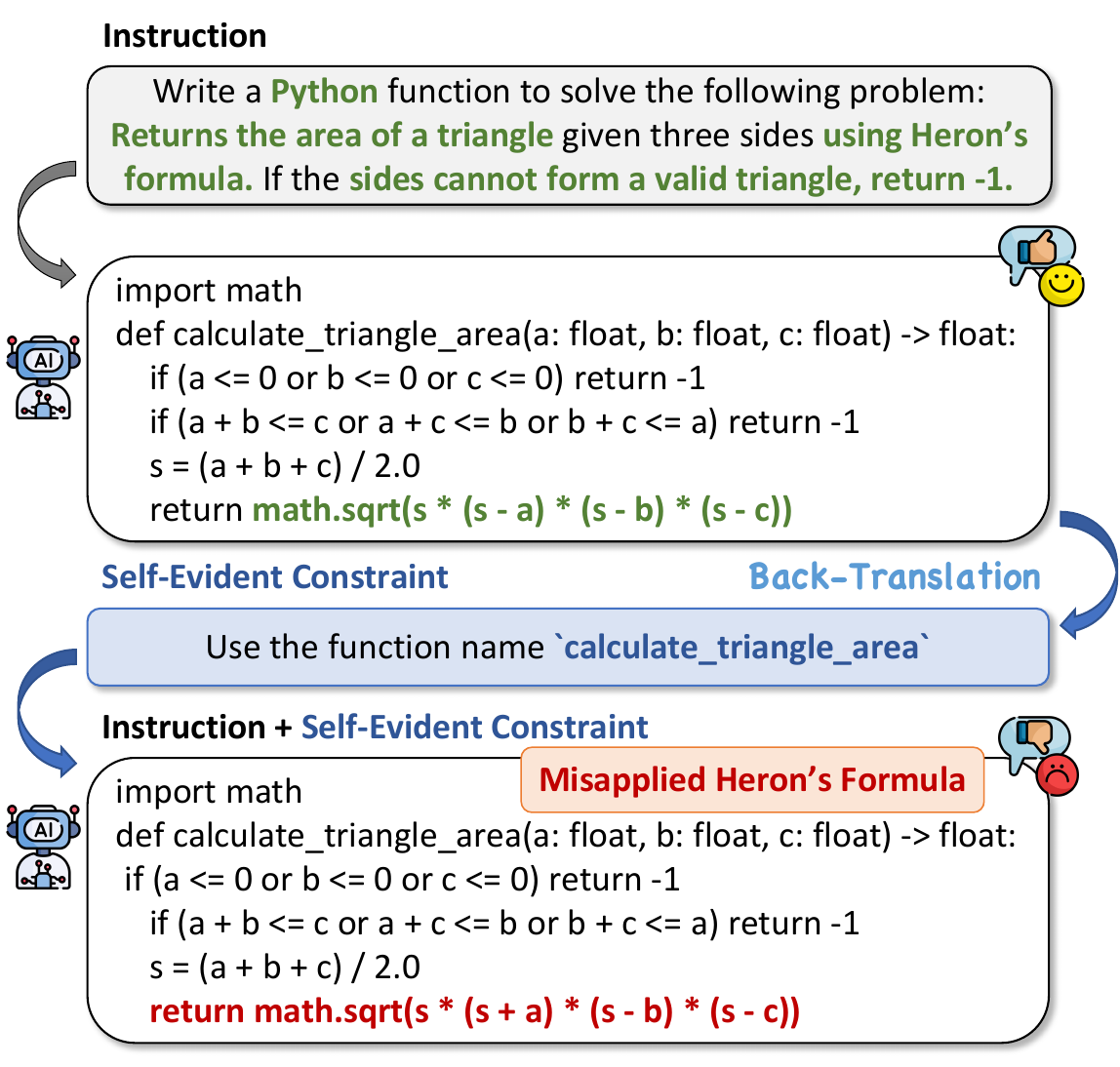} %pdf
    \caption{
    % Current models struggle to achieve the underlying goal of a task when faced with additional, yet reasonable, human constraints—a capability that is fundamental to any well-aligned model.
    LLMs may fail at a task merely by adding a self-evident constraint that is already being met in their originally successful, unconstrained outputs.
    % An illustration of our set-up. ...
    }
    \label{fig:fig1}
\end{figure}

% 指令遵循是什么 指令遵循的重要性 -> 促进了指令遵循评估数据集和相关技术的发展 -> 一个被忽视的问题。

Large Language Models (LLMs) have shown remarkable performance across diverse tasks and complex agentic workflows~\citep{gpt4, guo2025deepseek, li2024agentsurvey}.
One important factor behind this progress is instruction following, which allows human to align model behavior with their intent through explicit constraints and task specifications~\citep{IF-survey, sun2024conifer, qi2024constraint}.
Consequently, recent alignment pipelines and evaluation frameworks increasingly emphasize constraint satisfaction, driving the development of numerous instruction-following benchmarks~\citep{zhou2023instruction,qin2024infobench,cello} and targeted optimizations in advanced models (e.g., GPT-5~\citep{openai2025gpt5}, Qwen3~\citep{yang2025qwen3}, and GLM-4.6~\citep{zeng2025glm}).
However, while instruction following is crucial for alignment, its potential impact on the model’s general task-solving abilities remains underexplored. In particular, \textit{does adding self-evident constraints affect the model's task-solving performance?}

In this paper, we formalize this problem as the interference of instruction following with LLMs' task-solving capabilities, and define the underlying capability as the robustness of a model's task-solving abilities under self-evident constraints.
We introduce an automated evaluation framework and present the first systematic empirical study of this interference. 
% We formalize the capability underlying above question as the robustness of a model’s task-solving abilities under self-evident constraints.
% We introduce an automated evaluation framework and provide the first systematic empirical study of this interference of instruction following with task solving.
Specifically, we propose a metric, \textbf{\ourmetric}, which evaluates this interference by measuring how much task performance is sustained after self-evident constraints are imposed.
% However, task failures may arise from inherent capability limitations, such as the model being originally unable to solve the task, or imposed unreasonable constraints forcing the model into a reasoning path that exceeds its capabilities~\cite{fu2025scaling}.
% Hence, 
To isolate the interference of capability limitations, we utilize only instances that the model solves successfully without constraints and construct self-evident constraints by deriving them directly from its own successful outputs, ensuring that the model inherently possesses the capability to solve the task under these constraints.
We consider five common and widely used forms of constraints, including \textit{Method}, \textit{Length}, \textit{Style}, \textit{Keyword}, and \textit{Structure} constraints~\citep{zhou2023instruction}.
We evaluate LLMs of varying sizes and paradigms across Mathematics, Multi-hop QA, and Code.
Through extensive experiments, we uncover a paradoxical interference: 
% \textit{Adding constraints \textbf{degrades} general task-solving performance, even if these constraints are reasonable.}
\textit{adding even self-evident constraints significantly degrades general task-solving performance.}
Figure~\ref{fig:fig1} illustrates a concrete instance of this fragility, where a self-evident constraint derived from a model’s own successful solution leads to unexpected task failure.
These performance drops are substantial. Even advanced models such as Claude-Sonnet-4.5~\cite{claude45sonnet} show measurable losses, whereas many strong models with 30B to 70B parameters retain only around $65\%$ to $85\%$ of their original performance.
These results highlight a general lack of robustness in current models when executing tasks with constraints.
This gap is not fully captured by instruction-following and task-solving benchmarks, 
implying that benchmark success does not guarantee proficiency in realistic, constrained scenarios.

To investigate whether this paradoxical interference is prevalent, how it impacts model performance, and what its possible underlying reasons are, we conduct three comprehensive analyses.
\textbf{First}, we show that this performance degradation is consistent across experimental setups, supporting the reliability of \ourmetric.
We first confirm that the observed performance decline is not a mere artifact of instruction design, such as increased context length or structural changes. Furthermore, by analyzing constraint types and scaling effects, we find that performance degrades consistently across diverse constraint categories. In terms of constraint scaling, the performance exhibits a sharp initial decline with the addition of the first few constraints, followed by a plateau phase where the degradation stabilizes.
\textbf{Second}, we investigate the failure patterns and potential mechanisms behind this performance degradation.
At the behavioral level, we observe two dominant error modes: (1) \textit{Reasoning Error}, where the imposition of constraints causes the model to deviate from the correct reasoning trajectory; and (2) \textit{Output Specification Error}, where the model derives the correct core solution but fails to satisfy specific output or completeness requirements.
To investigate the underlying mechanisms,
we introduce a \textit{Constraint Attention Score}, measuring the proportion of attention focused on constraints during generation. 
Our comparison indicates that failed cases tend to exhibit higher constraint attention scores than successful ones, suggesting that the performance degradation may stem from an excessive attentional focus on constraint parts. 
\textbf{Finally}, we conduct a preliminary analysis of possible underlying reasons rooted in the training process. Our observations suggest that RL may simultaneously enhance both task performance and robustness under constraints. In contrast, while supervised fine-tuning on long chain-of-thought data boosts general performance, it appears significantly more susceptible to this degradation.

In summary, our contributions are threefold: 
(1) We identify the paradoxical interference between instruction following and task-solving, and formalize it as task robustness under constraints.
(2) We introduce \ourmetric to quantify this interference and demonstrate pervasive performance drops in current LLMs.
(3) We analyze this phenomenon from empirical validation to mechanistic interpretation, and provide practical insights for model alignment.
We encourage the community to emphasize task robustness under constraints to meet realistic, constrained scenarios. We also advise caution in instruction design, as excessive constraints may inadvertently degrade core task performance.
% (1) We formalize a principle of alignment, positing that a well-aligned model must maintain performance consistency across instructions that share the same underlying goal.
% (2) We propose \ourmetric, a novel and fair metric that 
% evaluates 
% this principle by turning a model's own successful outputs into explicit constraints.
% (3) We apply \ourmetric to conduct extensive experiments and assess the alignment of current models. We further show the distinct impacts of various alignment techniques and \ourmetric is a well predictor of human preferences for instruction-following.
% \section{\ourmetric: Evaluating Alignment}

% \section{\ourmetric: A Verifiable Metric For Robust and Genuine Alignment}
\section{\ourmetric}
\label{sec:methodology}

\begin{figure*}[t]
    \centering
    \includegraphics[width=1.0\linewidth]{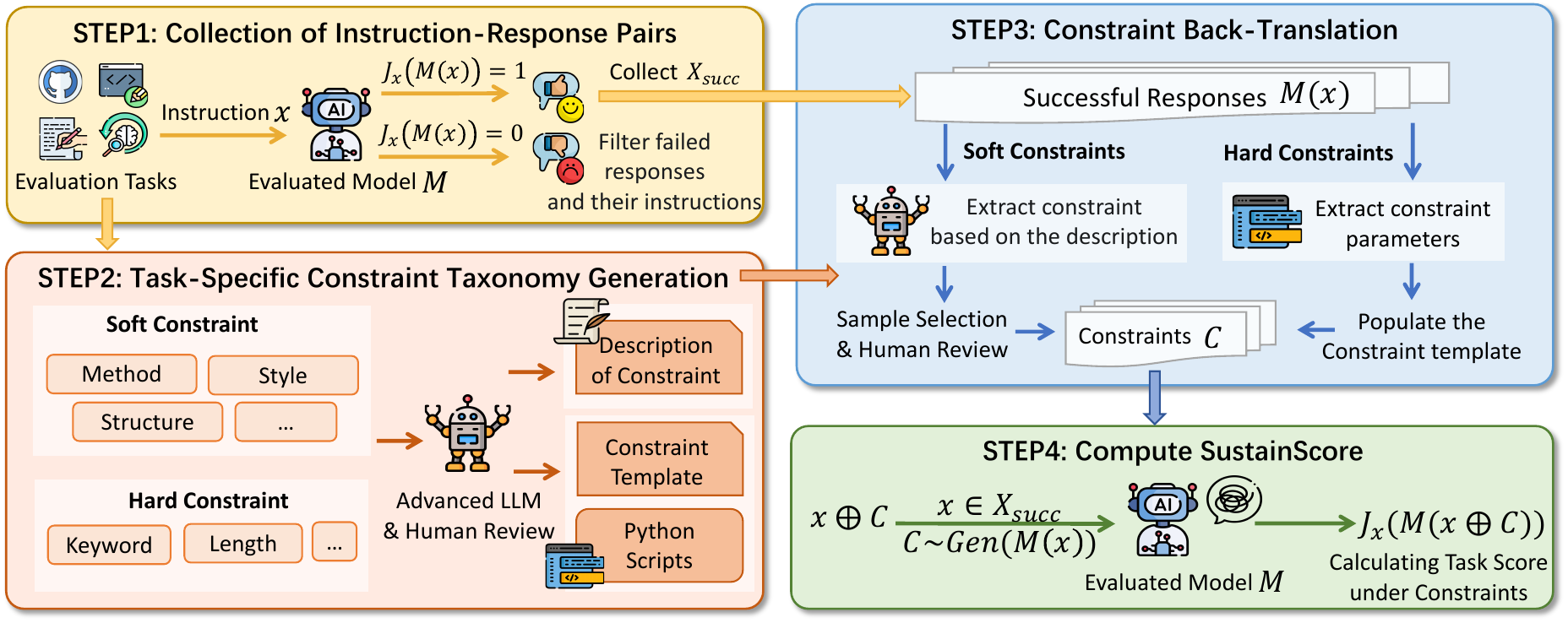} %pdf
    \caption{An overview of the evaluation framework for computing \ourmetric.
    }
    \label{fig:pipeline}
\end{figure*}

In this section, we first provide the formal definition of \ourmetric. This metric evaluates the impact of instruction following on task-solving capabilities, i.e., the robustness of task performance under constraints.
Then, we describe an automated evaluation framework to calculate \ourmetric based on existing datasets, allowing for scalability across domains without additional human effort.

\subsection{Definition}

Current benchmarks primarily evaluate instruction following by assessing whether a model's output strictly adheres to a set of explicit constraints~\citep{zhou2023instruction, qin2024infobench}.
This evaluation process can be formalized by considering a core task instruction $x$ and a set of constraints $\mathcal{C} = \{c_i\}_{i=1}^k$. The model $M$ receives the concatenated input as instruction $x \oplus \mathcal{C}$ (where $\oplus$ denotes concatenation) and produces an output $M(x \oplus \mathcal{C})$.
The instruction following success is determined by verifying whether $M(x \oplus \mathcal{C})$ satisfies all constraints in $\mathcal{C}$.

While this paradigm effectively measures explicit constraint adherence, it overlooks the impact of constraints on core task performance, i.e., whether the output $M(x \oplus \mathcal{C})$ can sustain the performance of $M(x)$.
To bridge this gap, we propose \ourmetric to evaluate task performance under constraints, serving as a measure of task robustness.
A critical challenge for measuring this robustness is ensuring that any observed performance drop is caused by the interference of instruction following, rather than the model's inability to solve the core task itself, or imposed constraints that render the task logically unsolvable.
Therefore, it is essential to ensure the solvability of the core task for model $M$ under the imposed constraints. Specifically, this requires satisfying two conditions: (1) Inherent Task Capability: the model must demonstrate the ability to solve the core task; and (2) Constraint Reasonableness: the imposed constraints must be \textit{self-evident}, meaning they align with a valid solution path and do not introduce conflicts that exceed the model's capabilities.

Accordingly, we formalize the calculation of \ourmetric as follows.
Let $J_x(y)=1$ indicate that an output $y$ correctly solves the core task $x$.
For a batch of instructions $\mathcal{X}$, we construct the valid subset $\mathcal{X}_{\text{succ}} = \{x \in \mathcal{X} \mid J_x(M(x))=1\}$, which contains only the tasks that the model can successfully solve without constraints.
Then, for each $x \in \mathcal{X}_{\text{succ}}$, we generate a set of \textit{self-evident} constraints $\mathcal{C}$ by deriving them from the successful output $M(x)$ (denoted as $\mathcal{C} \sim \text{Gen}(M(x))$), thereby ensuring that the model is inherently capable of satisfying both the task and the constraints.
We define \ourmetric for model $M$ as:
\begin{equation}
\label{eq:1}
\begin{split}
\ourmetric_M(\mathcal{X})
&= \mathbb{E}_{\,x \sim \mathcal{X}_{\mathrm{succ}},\, \mathcal{C} \sim \text{Gen}(M(x))} \\
&\quad \Big[ J_x\!\big(M(x \oplus \mathcal{C}\big) \Big]
\end{split}
\end{equation}
where $\mathbb{E}$ denotes the expected task performance averaged over the evaluation dataset.
This metric directly quantifies a model's ability to retain its task success under self-evident constraints.

\subsection{An automated evaluation framework} 

In this work, we empirically implement \ourmetric based on the formalization in Equation~\ref{eq:1}. Following prior work on instruction evaluation~\citep{zhou2023instruction,peng2025verif}, we consider five common constraint types in two categories. The first category, hard constraints, targets syntactic or lexical forms, including \textit{Length} (e.g., word or sentence count) and \textit{Keyword} (the inclusion or exclusion of specific terms). The second category, soft constraints, targets semantic content and organizational logic, including \textit{Method} (the core reasoning approach), \textit{Style} (the semantic tone), and \textit{Structure} (the output’s organization).

The full evaluation pipeline is illustrated in Figure~\ref{fig:pipeline}. 
For a given task dataset, the process begins by (1) collecting solvable instruction-response pairs while filtering out failures. Next, (2) we construct a task-specific constraint taxonomy based on the characteristics of the evaluated task. This taxonomy then guides (3) the constraint back-translation stage, where explicit constraints are derived from each successful response. Finally, (4) compute \ourmetric.
% these constraints are appended to the original instructions to form the final evaluation instructions. 
% The \ourmetric score quantifies the model's ability to maintain task performance on these new, constrained instructions, serving as a direct measure of its task-solving robustness under instruction following.
We detail the core stages below.

\paragraph{Collection of Instruction-Response Pairs.}
We generate model responses to the evaluation task dataset and filter out any failed instances. We collect only the pairs of instructions and responses where the model successfully solves the core task.

\paragraph{Task-Specific Constraint Taxonomy Generation.}
This process is tailored to the evaluation task domain (e.g., mathematics, code generation) and handles hard and soft constraints differently.
For hard constraints like \textit{Length} and \textit{Keyword}, which can be extracted by a rule-based approach, we use a two-step procedure. First, GPT-5~\citep{openai2025gpt5} generates a diverse taxonomy of sub-types and phrasing templates (e.g., for \textit{Length}, sub-types may include ``word count,'' ``number of paragraphs,'' or ``sentence count''). Second, for each sub-type, the LLM drafts a Python script to extract the needed concrete parameters (e.g., a specific word count) from a given model response. All templates and scripts are then manually verified for correctness.
For soft constraints such as \textit{Method}, \textit{Structure}, and \textit{Style}, where a rule-based approach is infeasible, we prompt GPT-5 to produce fine-grained, task-specific descriptions for each soft type (e.g., for math, the \textit{Method} description enumerates valid reasoning paths such as ``use a system of linear equations in two variables''). These descriptions serve as rubrics for the subsequent constraint generation step, enabling targeted extraction of complex, semantic constraints from model outputs.

\paragraph{Constraint Back-Translation.} 
With the taxonomy in place, we convert implicit properties of a successful response into an explicit constraint $c$, inspired by~\citet{qi2024constraint}.
For hard constraints, we execute the Python scripts on each successful output to extract concrete parameters and instantiate the corresponding template.
For soft constraints, we employ Qwen3-32B~\citep{yang2025qwen3} as a semantic extractor, which is guided by the taxonomy’s fine-grained descriptions and few-shot examples. The LLM analyzes a response to identify its specific method, style, or structure, and formulates this into a natural-language constraint.
To validate this process, we randomly sampled $200$ generated constraints and conducted a manual verification; results showed that $195$ ($97.5\%$) were consistent with the original response.
% These constraints collectively constitute the reasonable constraint set.
Because each constraint is grounded in behaviors the model has already demonstrated, the resulting set $\mathcal{C}$ is self-evident.

\paragraph{Compute \ourmetric.} 
Finally, we append five randomly ordered constraints (one per category) to the original instruction. \ourmetric is then computed as the model's task success rate on these new, constrained instructions, serving as a direct measure of its task-solving robustness under instruction following. See Appendix~\ref{sec:invariancescore_imple} for details.

\begin{table*}[t]
\centering
\scriptsize
\setlength{\tabcolsep}{2.8pt}
\renewcommand{\arraystretch}{1.12}
\begin{adjustbox}{max width=\textwidth}
\begin{tabular}{>{\raggedright\arraybackslash}p{3.4cm} c c c c c c c c c}
\toprule
\multirow{2}{*}{Model} 
& \multirow{2}{*}{IF} 
& \multicolumn{2}{c}{Mathematics} 
& \multicolumn{2}{c}{Multi-Hop QA} 
& \multicolumn{2}{c}{Code Generation} 
& \multicolumn{2}{c}{Overall AVG} \\
\cmidrule(lr){3-4} \cmidrule(lr){5-6} \cmidrule(lr){7-8} \cmidrule(lr){9-10}
&  & Accuracy & \ourmetric & Accuracy & \ourmetric & Accuracy & \ourmetric & Accuracy & \ourmetric \\
\midrule
Qwen2.5-1.5B-Instruct & \cellcolor[HTML]{FFFFFF}$ 53.8 $ & \cellcolor[HTML]{FFFFFF}$ 33.1 $ & \cellcolor[HTML]{FB7151}$ 65.3 $ & \cellcolor[HTML]{FFFFFF}$ 17.7 $ & \cellcolor[HTML]{FB7252}$ 65.6 $ & \cellcolor[HTML]{FFFFFF}$ 31.5 $ & \cellcolor[HTML]{F85F43}$ 61.6 $ & \cellcolor[HTML]{FFFFFF}$ 27.4 $ & \cellcolor[HTML]{FB6C4C}$ 64.1 $ \\
% DeepSeek-R1-Distill-Qwen-1.5B & \cellcolor[HTML]{FFFFFF}$ 49.9 $ & \cellcolor[HTML]{FFFFFF}$ 61.9 $ & \cellcolor[HTML]{FB7151}$ 65.4 $ & \cellcolor[HTML]{FFFFFF}$ 9.1 $ & \cellcolor[HTML]{67000D}$ 27.5 $ & \cellcolor[HTML]{FFFFFF}$ 44.9 $ & \cellcolor[HTML]{F34935}$ 57.4 $ & \cellcolor[HTML]{FFFFFF}$ 38.6 $ & \cellcolor[HTML]{DC2924}$ 50.1 $ \\
Qwen2.5-7B-Instruct & \cellcolor[HTML]{FFFFFF}$ 80.6 $ & \cellcolor[HTML]{FFFFFF}$ 67.3 $ & \cellcolor[HTML]{FDCEBB}$ 86.5 $ & \cellcolor[HTML]{FFFFFF}$ 18.8 $ & \cellcolor[HTML]{F44F39}$ 58.4 $ & \cellcolor[HTML]{FFFFFF}$ 65.3 $ & \cellcolor[HTML]{E12D26}$ 51.2 $ & \cellcolor[HTML]{FFFFFF}$ 50.5 $ & \cellcolor[HTML]{FB7151}$ 65.4 $ \\
DeepSeek-R1-Distill-Qwen-7B & \cellcolor[HTML]{FFFFFF}$ 71.6 $ & \cellcolor[HTML]{FFFFFF}$ 68.8 $ & \cellcolor[HTML]{FCB69B}$ 80.5 $ & \cellcolor[HTML]{FFFFFF}$ 13.0 $ & \cellcolor[HTML]{F44D38}$ 58.3 $ & \cellcolor[HTML]{FFFFFF}$ 63.4 $ & \cellcolor[HTML]{FDCEBB}$ 86.4 $ & \cellcolor[HTML]{FFFFFF}$ 48.4 $ & \cellcolor[HTML]{FC9D7F}$ 75.1 $ \\
Meta-Llama-3.1-8B-Instruct & \cellcolor[HTML]{FFFFFF}$ 83.5 $ & \cellcolor[HTML]{FFFFFF}$ 56.1 $ & \cellcolor[HTML]{FDD1BE}$ 87.2 $ & \cellcolor[HTML]{FFFFFF}$ 19.6 $ & \cellcolor[HTML]{FC8565}$ 69.8 $ & \cellcolor[HTML]{FFFFFF}$ 45.9 $ & \cellcolor[HTML]{F0402F}$ 55.6 $ & \cellcolor[HTML]{FFFFFF}$ 40.5 $ & \cellcolor[HTML]{FC8969}$ 70.8 $ \\
DeepSeek-R1-Distill-Llama-8B & \cellcolor[HTML]{FFFFFF}$ 69.9 $ & \cellcolor[HTML]{FFFFFF}$ 64.8 $ & \cellcolor[HTML]{FC9D7F}$ 75.2 $ & \cellcolor[HTML]{FFFFFF}$ 18.2 $ & \cellcolor[HTML]{F44F39}$ 58.4 $ & \cellcolor[HTML]{FFFFFF}$ 60.2 $ & \cellcolor[HTML]{FCA689}$ 77.2 $ & \cellcolor[HTML]{FFFFFF}$ 47.7 $ & \cellcolor[HTML]{FC8767}$ 70.3 $ \\
Qwen2.5-14B-Instruct & \cellcolor[HTML]{FFFFFF}$ 86.7 $ & \cellcolor[HTML]{FFFFFF}$ 73.3 $ & \cellcolor[HTML]{FEDECF}$ 90.3 $ & \cellcolor[HTML]{FFFFFF}$ 28.7 $ & \cellcolor[HTML]{FC8969}$ 70.6 $ & \cellcolor[HTML]{FFFFFF}$ 66.2 $ & \cellcolor[HTML]{BE151A}$ 42.8 $ & \cellcolor[HTML]{FFFFFF}$ 56.1 $ & \cellcolor[HTML]{FB7C5C}$ 67.9 $ \\
OpenReasoning-Nemotron-14B & \cellcolor[HTML]{FFFFFF}$ 66.4 $ & \cellcolor[HTML]{FFFFFF}$ 72.3 $ & \cellcolor[HTML]{FEE1D3}$ 91.1 $ & \cellcolor[HTML]{FFFFFF}$ 13.2 $ & \cellcolor[HTML]{FB7151}$ 65.4 $ & \cellcolor[HTML]{FFFFFF}$ 52.2 $ & \cellcolor[HTML]{F85F43}$ 61.6 $ & \cellcolor[HTML]{FFFFFF}$ 45.9 $ & \cellcolor[HTML]{FC9272}$ 72.7 $ \\
DeepSeek-R1-Distill-Qwen-14B & \cellcolor[HTML]{FFFFFF}$ 82.8 $ & \cellcolor[HTML]{FFFFFF}$ 76.7 $ & \cellcolor[HTML]{FED8C7}$ 88.8 $ & \cellcolor[HTML]{FFFFFF}$ 28.5 $ & \cellcolor[HTML]{FB7858}$ 67.0 $ & \cellcolor[HTML]{FFFFFF}$ 74.8 $ & \cellcolor[HTML]{FCBBA1}$ 81.7 $ & \cellcolor[HTML]{FFFFFF}$ 60.0 $ & \cellcolor[HTML]{FCAF93}$ 79.1 $ \\
QwQ-32B & \cellcolor[HTML]{FFFFFF}$ 90.7 $ & \cellcolor[HTML]{FFFFFF}$ 83.7 $ & \cellcolor[HTML]{FEDFD0}$ 90.6 $ & \cellcolor[HTML]{FFFFFF}$ 31.2 $ & \cellcolor[HTML]{FC9879}$ 74.1 $ & \cellcolor[HTML]{FFFFFF}$ 62.4 $ & \cellcolor[HTML]{DC2924}$ 50.0 $ & \cellcolor[HTML]{FFFFFF}$ 59.1 $ & \cellcolor[HTML]{FC8D6D}$ 71.5 $ \\
Qwen2.5-32B-Instruct & \cellcolor[HTML]{FFFFFF}$ 87.4 $ & \cellcolor[HTML]{FFFFFF}$ 73.5 $ & \cellcolor[HTML]{FEE5D8}$ 92.8 $ & \cellcolor[HTML]{FFFFFF}$ 26.8 $ & \cellcolor[HTML]{FCA285}$ 76.4 $ & \cellcolor[HTML]{FFFFFF}$ 72.3 $ & \cellcolor[HTML]{F14130}$ 55.9 $ & \cellcolor[HTML]{FFFFFF}$ 57.5 $ & \cellcolor[HTML]{FC9C7D}$ 75.0 $ \\
DeepSeek-R1-Distill-Qwen-32B & \cellcolor[HTML]{FFFFFF}$ 82.4 $ & \cellcolor[HTML]{FFFFFF}$ 76.1 $ & \cellcolor[HTML]{FEE3D7}$ 92.2 $ & \cellcolor[HTML]{FFFFFF}$ 32.8 $ & \cellcolor[HTML]{F96346}$ 62.4 $ & \cellcolor[HTML]{FFFFFF}$ 76.8 $ & \cellcolor[HTML]{FEE3D7}$ 92.1 $ & \cellcolor[HTML]{FFFFFF}$ 61.9 $ & \cellcolor[HTML]{FCBDA4}$ 82.3 $ \\
Qwen3-32B & \cellcolor[HTML]{FFFFFF}$ 90.2 $ & \cellcolor[HTML]{FFFFFF}$ 85.9 $ & \cellcolor[HTML]{FEE1D4}$ 91.3 $ & \cellcolor[HTML]{FFFFFF}$ 34.2 $ & \cellcolor[HTML]{FC9777}$ 73.9 $ & \cellcolor[HTML]{FFFFFF}$ 48.1 $ & \cellcolor[HTML]{FCAA8D}$ 78.1 $ & \cellcolor[HTML]{FFFFFF}$ 56.1 $ & \cellcolor[HTML]{FCB89E}$ 81.1 $ \\
GLM-Z1-32B & \cellcolor[HTML]{FFFFFF}$ 90.5 $ & \cellcolor[HTML]{FFFFFF}$ 85.7 $ & \cellcolor[HTML]{FEDCCD}$ 89.9 $ & \cellcolor[HTML]{FFFFFF}$ 35.3 $ & \cellcolor[HTML]{FB7656}$ 66.5 $ & \cellcolor[HTML]{FFFFFF}$ 43.3 $ & \cellcolor[HTML]{AB1016}$ 38.2 $ & \cellcolor[HTML]{FFFFFF}$ 54.8 $ & \cellcolor[HTML]{FB6E4E}$ 64.9 $ \\
Meta-Llama-3.1-70B-Instruct & \cellcolor[HTML]{FFFFFF}$ 90.7 $ & \cellcolor[HTML]{FFFFFF}$ 70.0 $ & \cellcolor[HTML]{FEE5D8}$ 92.9 $ & \cellcolor[HTML]{FFFFFF}$ 35.2 $ & \cellcolor[HTML]{FCAE92}$ 78.9 $ & \cellcolor[HTML]{FFFFFF}$ 62.4 $ & \cellcolor[HTML]{F85D42}$ 61.2 $ & \cellcolor[HTML]{FFFFFF}$ 55.9 $ & \cellcolor[HTML]{FCA98C}$ 77.7 $ \\
Qwen2.5-72B-Instruct & \cellcolor[HTML]{FFFFFF}$ 90.8 $ & \cellcolor[HTML]{FFFFFF}$ 74.5 $ & \cellcolor[HTML]{FEE0D2}$ 90.9 $ & \cellcolor[HTML]{FFFFFF}$ 32.2 $ & \cellcolor[HTML]{FCB398}$ 80.1 $ & \cellcolor[HTML]{FFFFFF}$ 72.9 $ & \cellcolor[HTML]{FB6C4C}$ 64.2 $ & \cellcolor[HTML]{FFFFFF}$ 59.9 $ & \cellcolor[HTML]{FCAB8F}$ 78.4 $ \\ 
% \midrule
Gemini-2.5-Flash & \cellcolor[HTML]{FFFFFF}$ 92.9 $ & \cellcolor[HTML]{FFFFFF}$ 72.7 $ & \cellcolor[HTML]{FDC6B0}$ 84.5 $ & \cellcolor[HTML]{FFFFFF}$ 42.9 $ & \cellcolor[HTML]{FC9373}$ 73.0 $ & \cellcolor[HTML]{FFFFFF}$ 76.1 $ & \cellcolor[HTML]{FDCEBB}$ 86.6 $ & \cellcolor[HTML]{FFFFFF}$ 63.9 $ & \cellcolor[HTML]{FCB99F}$ 81.4 $ \\
Grok-4-Fast & \cellcolor[HTML]{FFFFFF}$ 88.8 $ & \cellcolor[HTML]{FFFFFF}$ 68.5 $ & \cellcolor[HTML]{FCB69B}$ 80.7 $ & \cellcolor[HTML]{FFFFFF}$ 34.9 $ & \cellcolor[HTML]{FC997A}$ 74.5 $ & \cellcolor[HTML]{FFFFFF}$ 77.1 $ & \cellcolor[HTML]{FC9B7C}$ 74.8 $ & \cellcolor[HTML]{FFFFFF}$ 60.2 $ & \cellcolor[HTML]{FCA486}$ 76.7 $ \\
GPT-4.1-MINI & \cellcolor[HTML]{FFFFFF}$ 90.9 $ & \cellcolor[HTML]{FFFFFF}$ 78.9 $ & \cellcolor[HTML]{FEE8DD}$ 94.3 $ & \cellcolor[HTML]{FFFFFF}$ 40.2 $ & \cellcolor[HTML]{FCC3AB}$ 83.6 $ & \cellcolor[HTML]{FFFFFF}$ 77.1 $ & \cellcolor[HTML]{E02C26}$ 50.8 $ & \cellcolor[HTML]{FFFFFF}$ 65.4 $ & \cellcolor[HTML]{FCA285}$ 76.3 $ \\
% GLM-4.5 & \cellcolor[HTML]{FFFFFF}$ 92.9 $ & \cellcolor[HTML]{FFFFFF}$ 84.4 $ & \cellcolor[HTML]{FEE8DE}$ 94.1 $ & \cellcolor[HTML]{FFFFFF}$ 41.7 $ & \cellcolor[HTML]{F44F39}$ 55.6 $ & \cellcolor[HTML]{FFFFFF}$ 77.1 $ & \cellcolor[HTML]{67000D}$ 22.3 $ & \cellcolor[HTML]{FFFFFF}$ 67.7 $ & \cellcolor[HTML]{F6583E}$ 57.3 $ \\
Deepseek-V3.1 & \cellcolor[HTML]{FFFFFF}$ 92.5 $ & \cellcolor[HTML]{FFFFFF}$ 83.2 $ & \cellcolor[HTML]{FEE9DF}$ 94.8 $ & \cellcolor[HTML]{FFFFFF}$ 33.5 $ & \cellcolor[HTML]{FCA78B}$ 77.4 $ & \cellcolor[HTML]{FFFFFF}$ 77.1 $ & \cellcolor[HTML]{FEDCCD}$ 90.1 $ & \cellcolor[HTML]{FFFFFF}$ 64.6 $ & \cellcolor[HTML]{FDD2BF}$ 87.4 $ \\
Qwen3-235B-A22B-Instruct & \cellcolor[HTML]{FFFFFF}$ 94.0 $ & \cellcolor[HTML]{FFFFFF}$ 84.9 $ & \cellcolor[HTML]{FFF0E9}$ 97.8 $ & \cellcolor[HTML]{FFFFFF}$ 40.6 $ & \cellcolor[HTML]{FCC1A8}$ 83.2 $ & \cellcolor[HTML]{FFFFFF}$ 77.1 $ & \cellcolor[HTML]{FDC6B0}$ 84.7 $ & \cellcolor[HTML]{FFFFFF}$ 67.5 $ & \cellcolor[HTML]{FDD7C6}$ 88.5 $ \\
Claude-Sonnet-4-5 & \cellcolor[HTML]{FFFFFF}$ 93.5 $ & \cellcolor[HTML]{FFFFFF}$ 84.8 $ & \cellcolor[HTML]{FFEEE6}$ 96.7 $ & \cellcolor[HTML]{FFFFFF}$ 45.1 $ & \cellcolor[HTML]{FDC9B3}$ 85.0 $ & \cellcolor[HTML]{FFFFFF}$ 77.1 $ & \cellcolor[HTML]{FEE4D8}$ 92.6 $ & \cellcolor[HTML]{FFFFFF}$ 69.0 $ & \cellcolor[HTML]{FEE1D4}$ 91.4 $ \\
\bottomrule
\end{tabular}
\end{adjustbox}
\caption{
Model performance (\%) across IF (instruction following), Accuracy (task performance evaluated on the original datasets), and \ourmetric (robustness under constraints).
Overall AVG reports the macro-average.
The background color indicates the \hlneg{distance} from 100\% (perfect sustain), where darker shades indicate a more significant performance degradation.
}
% {Model performance (\%) across baseline tasks and our \ourmetric. Baseline performance is measured by instruction-following ability (IF) and task capability (Accuracy). AVG represents the average Accuracy and \ourmetric across all datasets. The background color indicates the \hlneg{distance} from 100\%, where darker shades indicate a more significant performance degradation.}
\label{tab:acc_by_suite_constraint_with_if}
\end{table*}

\section{Experiments}

\subsection{Experimental Setup}

\paragraph{Evaluation Benchmarks}
Our evaluation covers three core domains: mathematical reasoning, multi-hop question answering, and code generation.
(1) Mathematical reasoning: We use a total of $1,000$ samples, randomly drawing $250$ from each of the following four datasets: GSM8K~\citep{cobbe2021gsm8k}, SVAMP~\citep{SVAMP}, OlympiadBench~\citep{he2024olympiadbench}, and MATH500~\citep{lightman2023lets}.
(2) Multi-hop QA: For this reasoning task, we compiled a set of $1,200$ samples, with $400$ randomly drawn from each of HotpotQA~\citep{yang2018hotpotqa}, 2WikiMultiHop~\citep{xanh2020_2wikimultihop}, and Musique~\citep{trivedi2021musique}.
(3) Code generation: Our test set consists of $319$ samples, including the complete HumanEval benchmark~\citep{chen2021codex} ($164$ samples) and $150$ from MCEval~\citep{mceval} ($50$ each for C, C++, and Python).

\paragraph{Investigated Models}
Our evaluation covers a diverse spectrum of models, categorized into two primary tiers based on their capabilities and scale:
\noindent\textbf{(1) Flagship Models:} To benchmark the state-of-the-art (SOTA) models, we evaluate leading proprietary models including Claude-Sonnet-4.5~\citep{anthropic2024claude35sonnet}, Grok-4-Fast~\citep{xai2025-grok4-fast-model-card}, Gemini-2.5-Flash~\citep{comanici2025gemini}, and GPT-4.1-MINI~\citep{gpt4}. 
Crucially, we also include large open-weights models that are comparable to these closed systems, specifically Qwen3-235B-A22B-Instruct~\citep{yang2025qwen3},  and DeepSeek-V3.1~\citep{guo2025deepseek}.
\noindent\textbf{(2) Efficient Models:} Spanning the 1.5B to 72B range, this category targets standard-scale and resource-efficient deployment scenarios, including models from the Qwen3 series~\citep{yang2025qwen3}, Llama series~\citep{grattafiori2024llama}, distilled versions of the DeepSeek series~\citep{guo2025deepseek}, GLM series~\citep{zeng2025glm}, 
and their post-trained derivatives~\citep{hu2025openreasonerzeroopensourceapproach,chen2025acereason}.

\subsection{Experimental Results}

Our primary experimental results, presented in Table~\ref{tab:acc_by_suite_constraint_with_if}, evaluate model robustness of task performance under instruction following using \ourmetric.
We analyze \ourmetric alongside two reference metrics: the instruction following (IF) score, computed as the instruction-level accuracy from IFEval~\citep{zhou2023instruction} (the most widely used instruction-following benchmark), and the original unconstrained task accuracy (Accuracy), which reflects the model's inherent capability on the evaluated tasks. Analysis of our main findings reveals three critical insights:

(1) All models demonstrate suboptimal performance under \ourmetric.
Even the state-of-the-art model, Claude-Sonnet-4.5, retains only $85.0\%$ of its capability on Multi-Hop QA. 
The degradation is more drastic for other models. For instance, GLM-Z1-32B retains merely $38.2\%$ of its original performance on Code. 
In particular, many competitive models in the 32B to 72B parameter range retain only $64.9\%$ to $82.3\%$ of their performance on average. This widespread inability to maintain performance highlights the necessity of \ourmetric for quantifying this gap. 
% We suggest that the robustness of task performance under constraints should be a critical criterion in the final stages of model training and selection.

(2) High scores on instruction following and task-solving benchmarks do not guarantee high \ourmetric. 
An example is GPT-4.1-MINI, which achieves excellent instruction-following (IF: $90.9\%$) and strong task performance (Code Accuracy: $77.1\%$), yet its Code \ourmetric is only $50.8\%$.
Furthermore, even among strong models with IF scores exceeding $90.0\%$ and average Accuracy above $65.0\%$ (peaking at $69.0\%$), the Overall \ourmetric still spans a significant range, from $76.3\%$ to $91.4\%$.
This disconnection demonstrates that high task capability and strong instruction following do not necessarily translate into robust task execution under constraints.

(3) Code generation task exhibits the most severe degradation, with seven models retaining less than $60\%$ of their performance. 
% Furthermore, performance variance across models is substantially larger in code.
This likely stems from the strict nature of execution-based evaluation, coupled with the code's structural fragility, where constraints can disrupt global logic rather than just local phrasing.
Given the central role of code generation in agentic systems, this fragility highlights the importance of explicitly measuring robustness under constraints.

\section{Analysis}

% In this section, we deepen our understanding of this phenomenon through three comprehensive analyses.
In this section, we present three progressive analyses. First, we validate the generality of this paradoxical interference between instruction following and task solving (§~\ref{sec:ablation}). Second, we investigate how this interference leads to performance degradation from the perspectives of behavioral patterns and internal mechanisms (§~\ref{sec:error_analysis}). Finally, we conduct a preliminary analysis of the underlying causes rooted in the training process (§~\ref{sec:post_trainint}).
We hope our analysis provides a few fresh perspectives on constructing reliable models.

% conduct a preliminary exploration into how distinct post-training techniques affect this robustness

\subsection{Analysis on Prevalence of the Interference}
% \subsection{Analysis on \ourmetric Reliability}
\label{sec:ablation}

\begin{table}[t]
\centering
\small
\begin{adjustbox}{width=\linewidth}
\setlength{\tabcolsep}{4pt}
\renewcommand{\arraystretch}{1.15}
\begin{tabular}{lccccc}
\toprule
Model & $x_{\text{long}}$ & $Inst_0$ & $Inst_1$ & $Inst_2$ & $Inst_3$ \\
\midrule
Qwen2.5-7B-Instruct & \cellcolor[HTML]{FFFFFF}$ 92.7 $ & \cellcolor[HTML]{FDCCBA}$ 86.5 $ & \cellcolor[HTML]{FCB69B}$ 80.5 $ & \cellcolor[HTML]{FDCCB9}$ 86.3 $ & \cellcolor[HTML]{FDC5AF}$ 84.4 $ \\
Meta-Llama-3.1-8B-Instruct & \cellcolor[HTML]{FFFFFF}$ 92.6 $ & \cellcolor[HTML]{FDCFBD}$ 87.2 $ & \cellcolor[HTML]{FDCDBB}$ 86.8 $ & \cellcolor[HTML]{FDD9CC}$ 90.0 $ & \cellcolor[HTML]{FDD5C6}$ 88.9 $ \\
OpenReasoning-Nemotron-14B & \cellcolor[HTML]{FFFFFF}$ 95.7 $ & \cellcolor[HTML]{FEDED2}$ 91.1 $ & \cellcolor[HTML]{FEEBE4}$ 94.6 $ & \cellcolor[HTML]{FEE2D7}$ 92.3 $ & \cellcolor[HTML]{FEDBCE}$ 90.5 $ \\
Qwen2.5-14B-Instruct & \cellcolor[HTML]{FFFFFF}$ 93.2 $ & \cellcolor[HTML]{FEDBCD}$ 90.3 $ & \cellcolor[HTML]{FEDCCF}$ 90.6 $ & \cellcolor[HTML]{FEDDD0}$ 90.9 $ & \cellcolor[HTML]{FDD9CA}$ 89.8 $ \\
GLM-Z1-32B-0414 & \cellcolor[HTML]{FFFFFF}$ 97.9 $ & \cellcolor[HTML]{FDD9CB}$ 89.9 $ & \cellcolor[HTML]{FDD2C1}$ 87.9 $ & \cellcolor[HTML]{FDD5C5}$ 88.7 $ & \cellcolor[HTML]{FCB69B}$ 77.9 $ \\
Qwen3-32B & \cellcolor[HTML]{FFFFFF}$ 95.5 $ & \cellcolor[HTML]{FEDED2}$ 91.3 $ & \cellcolor[HTML]{FEDFD3}$ 91.4 $ & \cellcolor[HTML]{FEDED2}$ 91.2 $ & \cellcolor[HTML]{FEDED2}$ 91.3 $ \\
\bottomrule
\end{tabular}
\end{adjustbox}
\caption{Performance (\%) on instruction length and structure.
}
\label{tab:math_quintuple_prompts_para}
\end{table}

To ensure the reliability of our findings, we validate \ourmetric across a wide range of experimental setups.
(1) \textbf{Validity}: We first disentangle the impact of surface-level factors, confirming that the observed degradation is not merely an artifact of instruction length or structure.
(2) \textbf{Scalability}: We further conduct analyses over different constraint quantities and types to verify the robustness of \ourmetric under varying levels of complexity.
% We validate the universality of performance degradation across a wide range of experimental setups.
% (1) We first disentangle the impact of surface-level factors, confirming that the observed degradation is not merely an artifact of instruction length or structural format.
% (2) We further conduct analyses over different constraint quantities and constraint types to verify the stability of \ourmetric.

\paragraph{Instruction Length.}
A potential confounding variable in our main experiment is the increased token count of the constrained instruction ($x\oplus \mathcal{C}$) compared to the original ($x$). We conduct an experiment where we paraphrase $x$ to match the word count of constrained instruction $x\oplus \mathcal{C}$, creating a longer instruction $x_{\text{long}}$ that is semantically identical to $x$. The experiment is conducted on the math task, and results are detailed in Table~\ref{tab:math_quintuple_prompts_para}.
We find that model performance on $x_{\text{long}}$ shows only a negligible drop, suggesting that the significant performance degradation measured by \ourmetric is attributable to the interference of instruction following, not merely from the instruction length.

\paragraph{Instruction Structure.}
We also test whether our findings are sensitive to the structural arrangement of the task instruction and the constraint. 
We evaluate three alternatives to our default ``instruction + constraint'' format ($Inst_0$):
(1) Constraint-First Priming ($Inst_1$): The constraint is placed before the instruction to test for order-dependence effects.
(2) Task Prioritization ($Inst_2$): A verbose template that explicitly instructs the model to prioritize the correct solution first and the constraint second.
(3) Step-by-Step Guidance ($Inst_3$): A template that instructs the model to think first and then write an answer that meets the constraint. 
% Details in Appendix~\ref{sec:invariancescore_imple}.
Table~\ref{tab:math_quintuple_prompts_para} shows that the performance degradation persists across these structural variations.
% While minor performance fluctuations exist across the different formats, the \ourmetric for any given model remains in a narrow range. 
No single template alleviates the poor performance. This confirms that the robustness failure captured by \ourmetric is an intrinsic property of the model's capability, rather than an artifact of instruction design. 
% Details in Appendix~\ref{sec:app_constraint}.

\paragraph{Constraint Quantity.}
\label{sec:constraint_number}

\begin{figure}[t]
    \centering
    \includegraphics[width=1.0\linewidth]{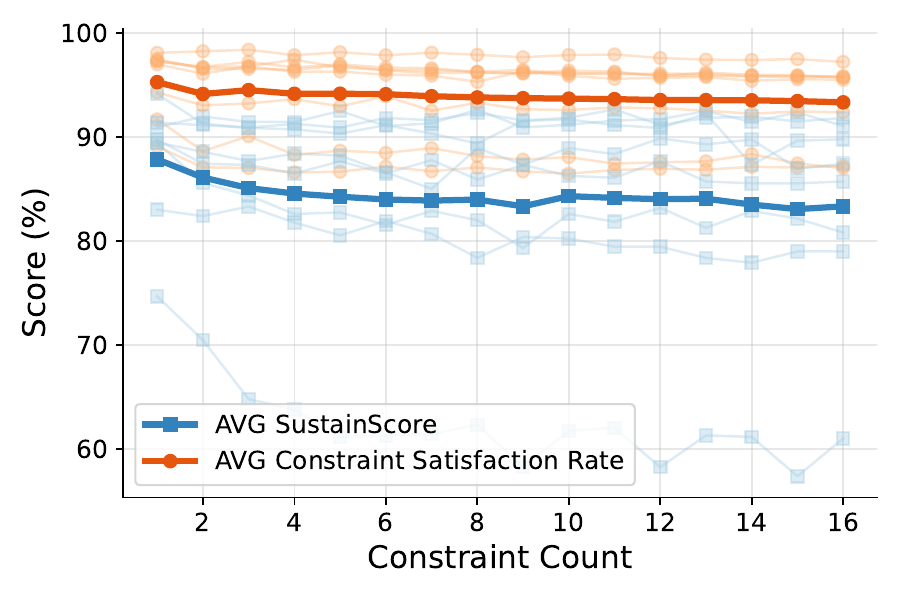} %pdf
    \caption{Comparison of \ourmetric and Constraint Satisfaction Rate as the number of constraints increases. 
    % Bold lines show the average across all models, while faint lines represent individual model performance.
    }
    \label{fig:scaling_up}
\end{figure}

To understand \ourmetric's scalability to the number of constraints, we conducted a constraint quantity analysis on the Math task across eight representative models using \textit{Keyword}. We increase the number of required keywords from $1$ to $16$, as shown in Figure~\ref{fig:scaling_up}.
The most important finding is the significant gap between a model's ability to follow instructions and its ability to maintain core task performance. While models consistently demonstrate a near-perfect ability to adhere to the constraints (orange line, averaging over $94\%$ satisfaction), their \ourmetric, which measures core task success, is significantly lower (blue line, averaging $84\%$). 
This gap demonstrates that existing instruction-following benchmarks, which primarily measure constraint satisfaction, 
fail to capture the hidden cost of constraints, treating format adherence as success even when the core task logic is broken.
Furthermore, the \ourmetric curve exhibits a sharp initial decline as constraints are added. Specifically, the most informative degradation occurs within the first $5$ constraints, after which the curve flattens, and the score largely stabilizes. This finding underscores the efficiency and practicality of \ourmetric, suggesting that an evaluation using just $5$ constraints is sufficient to capture the model's robustness.

\paragraph{Constraint Types.}
\label{sec:constraint_type}
To investigate the impact of different constraint categories on \ourmetric, we modified our standard procedure. We applied only a single constraint from each of the five categories in isolation, allowing us to measure how models respond to different constraint types. The results, averaged across all ``efficient models'' in our main experiment, are presented in Table~\ref{tab:constraint_type}.
First, performance is largely consistent across constraint types on average. The average performance gap between the best and worst-performing types is small ($7.8$), indicating that \ourmetric is not overly sensitive to the specific choice of constraint type. In practice, this flexibility allows the selection of constraint types that are easiest to implement without significantly skewing the final evaluation.
Second, the Code domain exhibits higher sensitivity to \textit{Keyword} ($70.0$). This aligns with the structural fragility discussed earlier: restricting specific keywords in code may risk disrupting global logic and execution flow, thereby amplifying the difficulty of the task.

\subsection{Analysis on Underlying Mechanisms}
\label{sec:error_analysis}

\begin{table}[t]
\centering
\small
\begin{adjustbox}{width=\linewidth}
\setlength{\tabcolsep}{4pt}
\begin{tabular}{lcccccc}
\toprule
Task & Length & Keyword & Style & Method & Structure & $\Delta$ \\
\midrule
Mathematics & $87.5$ & $89.0$ & $89.6$ & $90.4$ & $90.3$ & $2.9$ \\
Multi-Hop QA & $70.1$ & $73.2$ & $74.2$ & $74.9$ & $77.4$ & $7.3$ \\
Code Generation & $73.9$ & $70.0$ & $83.0$ & $86.9$ & $87.4$ & $17.4$ \\ \midrule
AVG & $77.2$ & $77.4$ & $82.3$ & $84.1$ & $85.0$ & $7.8$ \\
\bottomrule
\end{tabular}
\end{adjustbox}
\caption{\ourmetric on different constraint types. $\Delta$ indicates the max--min gap across constraint types.}
\label{tab:constraint_type}
\end{table}

To investigate the causes of model failure \textbf{at the behavioral level}, we manually inspected $50$ failed cases across domains and identified two dominant error types:
(1) \textit{Reasoning Error}: The model fails to establish a correct reasoning chain, leading to incorrect mathematical derivations, erroneous multi-hop inferences, or flawed algorithmic decisions in code. 
(2) \textit{Output Specification Error}: The model derives the correct core solution but fails to satisfy specific output or completeness requirements. Failures range from formatting mismatches (e.g., missing units or pattern deviations) to incomplete implementations (e.g., missing boundary cases), despite having the capability to handle them.
% (2) \textit{Output Specification Error}: The model derives the correct solution but fails to satisfy the required output or completeness conditions.  
% In Math and QA, this includes format mismatches such as missing units or deviations from required output patterns.  
% In Code, this often appears as missing task-required boundary cases or incomplete implementations, despite the model being capable of handling them in the unconstrained setting.
We further analyze the distribution of these two error types across evaluated models on math.
% , including $4$ direct-answer models and $4$ chain-of-thought (CoT) reasoning models. 
We observe that direct-answer models are more prone to Reasoning Errors, whereas CoT models frequently suffer from Output Specification Errors. 
This aligns with prior findings that while CoT enhances reasoning, it often exhibits weaker instruction following capabilities~\citep{li2025thinking}.
Examples and distributional statistics are provided in the Appendix~\ref{app:error_analysis}.

% \subsection{Constraint Attention Analysis}
% \label{sec:attention}

\begin{figure}[t]
    \centering
    \includegraphics[width=1.0\linewidth]{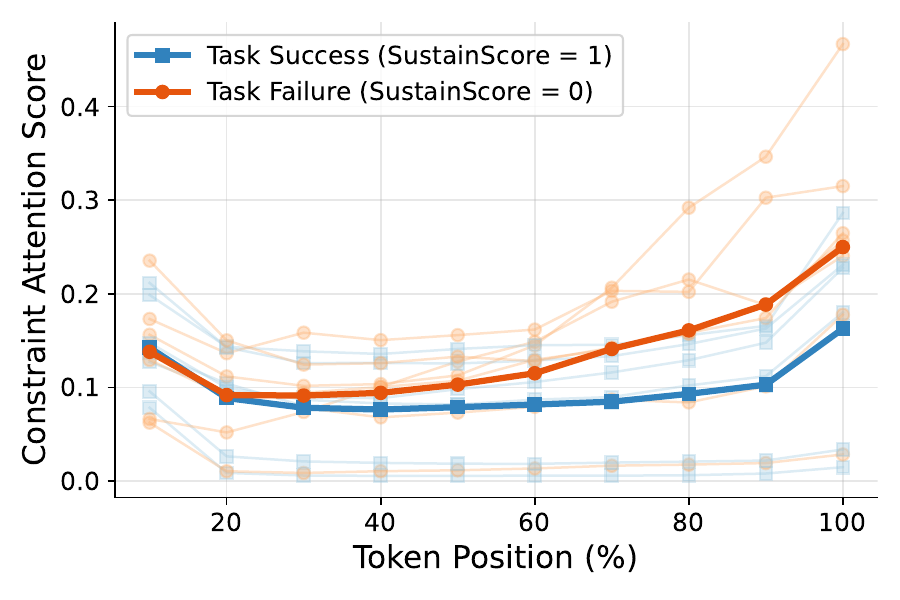} %pdf
    \caption{Comparison of constraint attention scores between successful and failed generations.
    }
    \label{fig:attention}
\end{figure}

To investigate these failures \textbf{at the mechanism level}, 
we conduct an attention-based analysis that tracks the model’s attention on constraint tokens throughout response generation. We define \textit{Constraint Attention Score} to quantify how prominently constraint tokens appear among the tokens that carry most of the model’s attention mass.
Specifically, at each generation step and for each attention head, we identify the smallest set of tokens whose cumulative attention weight reaches $50\%$ proportion of the total attention mass. We then compute the fraction of these high-attention tokens that belong to the constraint and normalize it by the proportion of constraint tokens in the entire context, yielding a head-level constraint attention score.
To obtain a token-level score, we average this head-level score over all attention heads in the last $4$ transformer layers at the same generation step. 
% A higher constraint attention score therefore indicates that constraint tokens occupy a larger share of the model’s highest-attention focus during generation. 
More details are provided in Appendix~\ref{app:attention}.

\begin{figure*}[t]
    \centering
    \includegraphics[width=1.0\linewidth]{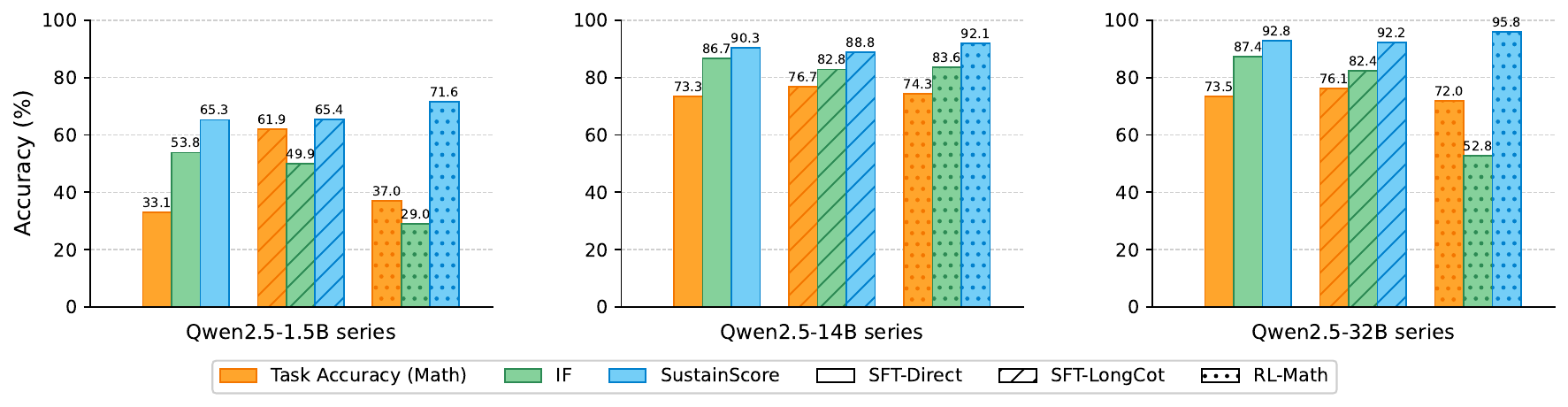} %pdf
    \caption{\ourmetric for different post-training strategies within the same model family on the Math task.
    }
    \label{fig:rl-sft}
\end{figure*}
% \section{Applications}

We compare the constraint attention scores of successful versus failed generations on math tasks across seven representative models. As shown in Figure~\ref{fig:attention}, failed cases consistently exhibit higher constraint attention scores than successful ones across most decoding steps. This gap becomes increasingly pronounced toward the later stages of generation: while both curves rise as decoding proceeds, the failure curve grows much more sharply, indicating that unsuccessful generations allocate a disproportionate amount of attention to constraint tokens near the end of the sequence. This suggests that excessive attention to constraints may interfere with the model's reasoning, offering a plausible explanation for the observed performance decline.

% \subsection{Application of \ourmetric}
\subsection{Analysis on Post-Training Factors}
\label{sec:post_trainint}

A critical application of any evaluation metric is to diagnose the shortcomings of existing techniques, thereby guiding future model development~\cite{zheng2023judgingllmasajudgemtbenchchatbot, zhou2023instruction}. 
Hence, we employ \ourmetric to explore the potential impact of post-training paradigms.
We compare models within the Qwen2.5 ecosystem that are built upon the same base architecture but differentiated by three alignment strategies:
(1) SFT-Direct: Supervised Fine-Tuning with direct answers, the standard approach for most instruction-tuned models (e.g., the Qwen2.5-Instruct series).
(2) SFT-LongCoT: Supervised Fine-Tuning on long CoT data, such as the DeepSeek-R1-Distill series~\citep{guo2025deepseek}.
(3) RL-Math: Reinforcement Learning specifically for mathematical reasoning, such as the Open-Reasoner-Zero series~\citep{hu2025openreasonerzeroopensourceapproach}.
As shown in Figure~\ref{fig:rl-sft}, our preliminary observations suggest that RL-based models tend to maintain better robustness alongside task performance. Conversely, SFT-LongCoT, while strong in task capability, appears more susceptible to performance degradation under constraints. This robustness gap is not fully reflected by standard metrics like IF, which focus on constraint satisfaction but overlook the preservation of core task capabilities, highlighting the complementary value of \ourmetric.

\section{Related Work}

\paragraph{Instruction Following.}
Instruction following constitutes a cornerstone of Large Language Model (LLM) alignment. Current instruction-following benchmarks typically assess whether a model’s response satisfies given constraints in a binary manner~\citep{zhou2023instruction, qin2024infobench, cello, qi2025agentif, complexbench, fu2025scaling}. However, these approaches face two significant limitations. First, constructing such benchmarks often requires extensive manual annotation and verification~\citep{qi2025agentif, qin2024infobench}, thereby limiting their scalability. Second, assessing adherence to complex constraints, such as stylistic conformity, frequently relies on LLM-based automatic evaluation~\citep{jiang-etal-2024-followbench}, which inevitably introduces noise and potential bias~\citep{ye2025justice}. More critically, effective alignment should entail not merely adhering to superficial instructions but also successfully accomplishing the core task. To this end, we propose \ourmetric, designed to capture a crucial dimension of alignment overlooked by prior instruction following works: the capability to maintain core task performance under self-evident constraints.

\paragraph{Prompt Sensitivity.}
Prompt sensitivity is commonly defined as the phenomenon where semantically equivalent reformulations of an instruction lead to substantial variations in model behavior or task performance~\citep{ps_chatterjee2024posix, ps_polo2024efficient, ps_mizrahi2024state, ps_voronov2024mind}. 
Most prior work studies this effect through surface-level perturbations, such as changes in punctuation, formatting, or phrasing, or through paraphrasing that preserves task semantics~\citep{razavi2025benchmarking, ps_mizrahi2024state}.
In contrast, we move beyond surface-level sensitivity to evaluate robustness under functional constraints (e.g., formatting or tool usage), which are common in real-world use.
These constraints preserve the task goal but alter how it should be carried out. 
We quantify a critical robustness gap: the inability to sustain performance under constraints, a dimension not captured by classical prompt sensitivity benchmarks.

% In contrast, we move beyond surface-level sensitivity to evaluate robustness under functional constraints. 
% In real-world use, prompts differ not only in wording but also in requirements such as formatting rules, content restrictions, or tool usage. These constraints preserve the task goal but alter how it should be carried out. Our work quantifies a critical robustness gap: the inability to sustain task performance when constraints are imposed, a dimension not captured by classical prompt sensitivity benchmarks.
\section{Conclusion}

% \looseness=-1
In this paper, we reveal a counterintuitive phenomenon: instruction following can paradoxically interfere with LLMs’ task-solving capability.
We introduce \ourmetric to quantify this interference. Our results reveal that this interference is severe and prevalent. Additionally, attention-based analyses suggest that the performance drop may stem from an excessive focus on constraints.
We further provide preliminary insights into how different post-training paradigms influence this interference.
We caution that excessive instructions may degrade model performance, and encourage future work to adopt \ourmetric to foster the development of more capable and reliable models.

 % We hope \ourmetric will guide future work in building compliant yet robust models.
 
% In this paper, we present the first systematic study on how instruction following impacts task performance. We formalize this capability as robustness under constraints and propose \ourmetric, an automated and scalable metric to measure it. Our evaluations reveal that even LLMs that excel on instruction-following and task benchmarks exhibit significant weakness under \ourmetric, highlighting a critical alignment dimension overlooked by prior work. We further demonstrate its utility in assessing alignment strategies. We encourage the community to adopt \ourmetric to foster the development of more capable and reliable models by strengthening this essential aspect of alignment.

\section*{Limitations}

The main limitations lie in the implementation of \ourmetric: (1) Our validation is confined to mathematical reasoning, code generation, and multi-hop QA. While effective in these domains, the generalizability of \ourmetric across more diverse tasks remains to be explored. While our evaluation is fully automated and easy to scale, we encourage the community to apply our metric to a broader range of scenarios.
(2) Our experiments are conducted in English. We recognize that linguistic differences could impact the results and welcome further investigation across different languages.
In conclusion, we acknowledge these limitations not as shortcomings, but as important directions for future work. We believe that \ourmetric is a significant contribution, and we look forward to future research that builds more reliable models for real-world scenarios.

\section*{Ethical Statements}
(1) Datasets: Our research utilizes widely-used evaluation datasets, and we strictly adhere to their licensing terms. We believe the original data is properly anonymized, and our work does not introduce any new sensitive information.
(2) Methodology Risks: Our proposed metric, \ourmetric, relies on an LLM generator for soft constraint creation. It therefore inherits the inherent risks of the generator model. However, our approach does not introduce additional risks. A manual review of $200$ randomly sampled instances confirmed the absence of sensitive information.
(3) AI Assistance: We used AI assistants (ChatGPT and Gemini) for language refinement and editing of this manuscript.

\bibliography{custom}

\newpage
\clearpage
\appendix
\section*{Appendices}
\section{\ourmetric  Details}
\label{sec:invariancescore_imple}
% inference 细节：prompt，vllm参数

Tables~\ref{tab:approach_instruction_prompt}, \ref{tab:format_instruction_prompt}, and \ref{tab:style_instruction_prompt} list the prompts for the \textit{Method}, \textit{Structure}, and \textit{Style} constraint types in \ourmetric, while Table~\ref{tab:instruction_length} provides those used for the instruction length analysis (§~\ref{sec:ablation}). For constraint generation, we employ Qwen3-32B~\cite{yang2025qwen3} with a maximum token length of $32{,}000$ and temperature set to $0$.
Finally, across all evaluations, we use a maximum token length of $8{,}192$ and temperature $0$ to ensure reproducibility.

\section{Detailed Results}
\label{sec:detail results}
% Prompt
% 细粒度的实验结果（figure3 figure4的）
% Code的结果
% Table~\ref{tab:code} shows the results of \ourmetric on code generation, which are consistent with our findings in math and multi-hop QA.
% Table~\ref{tab:math-if-align} shows the detailed results of the analysis of alignment techniques.

\subsection{Error Analysis}
\label{app:error_analysis}
% 错误case和distribution：fig:error_type

\begin{figure}[t]
    \centering
    \includegraphics[width=1.0\linewidth]{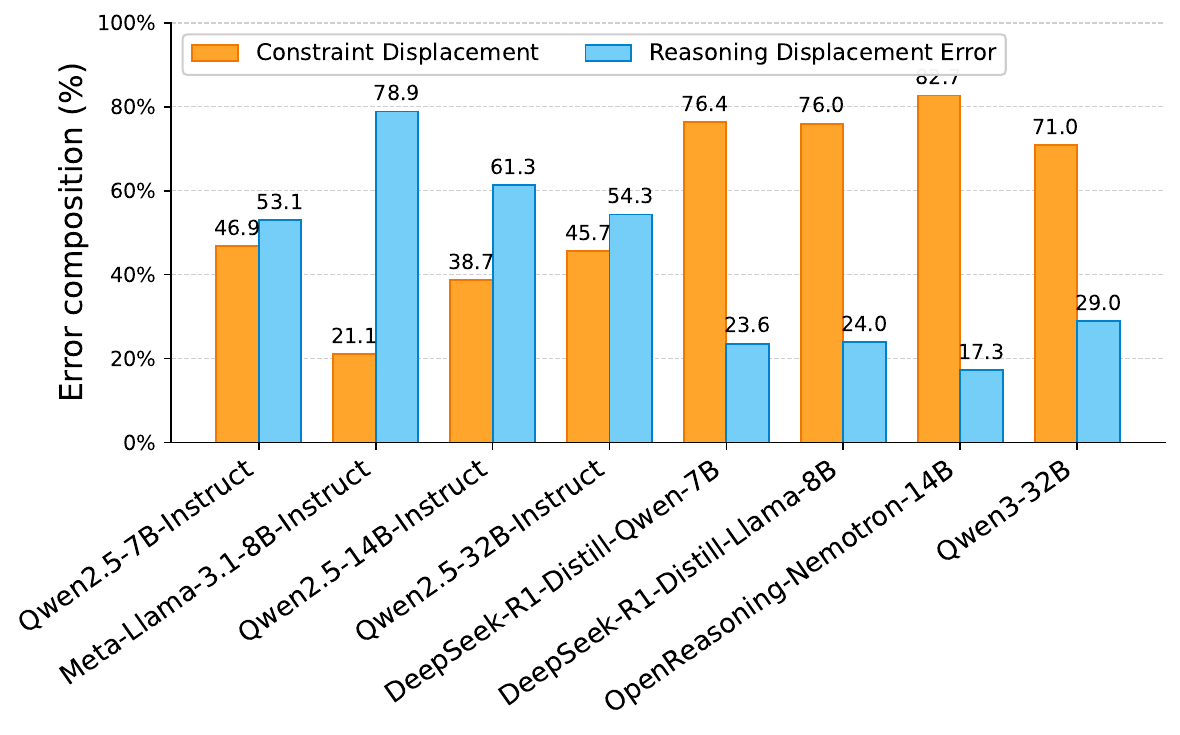} %pdf
    \caption{\ourmetric on different constraint types on Math. The score for each type represents the average performance across all models evaluated in our study.
    }
    \label{fig:error_type}
\end{figure}

Tables~\ref{tab:error_case},\ref{tab:code_case}, provide representative examples of the two error types.
Figure~\ref{fig:error_type} reports their distribution across eight models on math tasks, including $4$ direct-answer models and $4$ chain-of-thought reasoning models.
Because many output specification errors stem from formatting mismatches that cause exact-match or regex-based metrics to fail despite correct underlying solutions, we use an LLM-as-judge to determine semantic correctness.
Outputs judged as correct by the LLM are labeled as output specification errors; otherwise, they are considered reasoning errors.

\subsection{Constraint Attention Score}
\label{app:attention}
% 公式表达的结果
\paragraph{Constraint Attention Score.}
Let $a_{t,l,h}(i)$ denote the attention weight assigned to the $i$-th token in the context at generation step $t$, layer $l$, and attention head $h$, where $i=1,\dots,N_t$ and $N_t$ is the number of tokens in the context.
Let $S_2 \subseteq \{1,\dots,N_t\}$ be the set of constraint tokens.

\paragraph{Top-$p$ attention set.}
Given a proportion threshold $p=0.5$, we define the minimal top-$p$ attention set as
\begin{equation}
\mathcal{T}^{(p)}_{t,l,h}
=
\arg\min_{S \subseteq \{1,\dots,N_t\}}
\left\{
\sum_{i \in S} a_{t,l,h}(i)
\;\ge\;
p \sum_{j=1}^{N_t} a_{t,l,h}(j)
\right\}.
\end{equation}

\paragraph{Head-level score.}
We define the head-level constraint attention score as
\begin{equation}
c^{\mathrm{global}}_{t,l,h}
=
\frac{|\mathcal{T}^{(p)}_{t,l,h} \cap S_2|}{|\mathcal{T}^{(p)}_{t,l,h}|}
\;\Big/\;
\frac{|S_2|}{N_t}
\end{equation}

This score measures how much constraint tokens are over-represented in the
high-attention region relative to their random baseline.

\paragraph{Token-level score.}
Let $\mathcal{L}_K$ denote the set of the last $K$ transformer layers
(we use $K=4$), and let $H_l$ be the number of attention heads in layer $l$.
The token-level constraint attention score at generation step $t$ is defined as
\begin{equation}
c^{\mathrm{global}}_{t}
=
\frac{1}{K}
\sum_{l \in \mathcal{L}_K}
\left(
\frac{1}{H_l}
\sum_{h=1}^{H_l}
c^{\mathrm{global}}_{t,l,h}
\right).
\end{equation}

\subsection{Results on Post-Training Analysis}
Table~\ref{tab:math-if-align} presents the results discussed in Section~\ref{sec:post_trainint}. 
We additionally include the LLaMA series in the table below. These models exhibit the same trends, further supporting our observations.

\begin{table*}[t]
\centering
\small
\setlength{\tabcolsep}{6pt}
\begin{tabular}{l l c c c}
\toprule
Series & Model & Task Accuracy & \ourmetric & IF \\
\midrule
\multicolumn{5}{l}{\textbf{Qwen2.5-1.5B series}}\\
SFT-Direct   & Qwen2.5-1.5B-Instruct                 & 33.1 & 65.3 & 53.8 \\
SFT-LongCot  & DeepSeek-R1-Distill-Qwen-1.5B         & 61.9 & 65.4 & 49.9 \\
RL-Math    & Open-Reasoner-Zero-1.5B               & 37.0 & 71.6 & 29.0 \\
\midrule
\multicolumn{5}{l}{\textbf{Qwen2.5-14B series}}\\
SFT-Direct   & Qwen2.5-14B-Instruct                  & 73.3 & 90.3 & 86.7 \\
SFT-LongCot  & DeepSeek-R1-Distill-Qwen-14B          & 76.7 & 88.8 & 82.8 \\
RL-Math    & AceReason-Nemotron-14B                & 74.3 & 92.1 & 83.6 \\
\midrule
\multicolumn{5}{l}{\textbf{Qwen2.5-32B series}}\\
SFT-Direct   & Qwen2.5-32B-Instruct                  & 73.5 & 92.8 & 87.4 \\
SFT-LongCot  & DeepSeek-R1-Distill-Qwen-32B          & 76.1 & 92.2 & 82.4 \\
RL-Math    & Open-Reasoner-Zero-32B                & 72.0 & 95.8 & 52.8 \\
\midrule
\multicolumn{5}{l}{\textbf{Llama3.1-8B series}}\\
SFT-Direct   & Llama3.1-8B-Instruct                  & 56.1 & 87.2 & 83.5 \\
SFT-LongCot  & DeepSeek-R1-Distill-Llama-8B          & 64.8 & 75.2 & 69.9 \\
RL-Math    & Llama3.1-8B-Grpo-Reasoning                & 53.3 & 	88.0 & 82.6 \\
\bottomrule
\end{tabular}
\caption{Detailed results of the analysis of alignment techniques}
\label{tab:math-if-align}
\end{table*}

\begin{table*}[t]
\centering
\small
\setlength{\tabcolsep}{6pt}
\renewcommand{\arraystretch}{1.12}
\begin{tabularx}{\linewidth}{>{\raggedright\arraybackslash}X}
\toprule

You will be given a model response. Extract the core \emph{answering approach} it uses and rewrite it as a single, imperative instruction that can be attached to future prompts.\\[6pt]

\textbf{Instructions}\\[-2pt]
\textbullet\ Return \textbf{JSON only} with the key \texttt{"approach\_instruction"}.\\
\textbullet\ The value must be \textbf{one sentence} (\(\leq\) 25 words) starting with: \textbf{``Your answer should follow this approach: ''}.\\
\textbullet\ Summarize the \textbf{strategy}, not problem-specific facts or solutions.\\
\textbullet\ \textbf{No explanations, no extra keys, no text outside JSON.}\\
\textbullet\ If the approach is implicit, infer the most plausible high-level method.\\[6pt]

\textbf{Example}\\[-2pt]
\emph{Input Response:} \texttt{``Step 1: Kim's Contribution: Kim raised \$320 more than Alexandra, who raised \$430. \$430 + \$320 = \$750. Step 2: Maryam's Contribution: Maryam raised \$400 more than Sarah, who raised \$300. \$300 + \$400 = \$700. Step 3: Total Amount: We add: \$430 (Alexandra) + \$750 (Kim) + \$300 (Sarah) + \$700 (Maryam) = \$2180. Final Answer: 2180.''}\\[2pt]
\emph{Output:} \texttt{\{"approach\_instruction": "Your answer should follow this approach: compute each person's amount from provided relations, then sum all contributions; show step-by-step arithmetic leading to the final total."\}}\\[6pt]
\textbf{Your Turn}\\[-2pt]
\emph{Input Response:} ``\{response\}''\\
\emph{Output (Please directly return JSON only):}\\
\bottomrule
\end{tabularx}
\caption{Prompt for extracting Method constraints}
\label{tab:approach_instruction_prompt}
\end{table*}

\begin{table*}[t]
\centering
\small
\setlength{\tabcolsep}{6pt}
\renewcommand{\arraystretch}{1.12}
\begin{tabularx}{\linewidth}{>{\raggedright\arraybackslash}X}
\toprule

You will be given a model response. Extract the core \emph{language style} it uses and rewrite it as a single, imperative instruction that can be attached to future prompts.\\[6pt]

\textbf{Instructions}\\[-2pt]
\textbullet\ Return \textbf{JSON only} with the key \texttt{"style\_instruction"}.\\
\textbullet\ The value must be \textbf{one sentence} (\(\leq\) 25 words) starting with: \textbf{``Your answer should adopt this style: ''}.\\
\textbullet\ Summarize the \textbf{tone / style}, not content or problem-specific facts.\\
\textbullet\ \textbf{No explanations, no extra keys, no text outside JSON.}\\
\textbullet\ If the style is implicit, infer the most plausible high-level style.\\[6pt]

\textbf{Example}\\[-2pt]
\emph{Input Response:} \texttt{``Don't worry, you can think of it step by step. Step 1: Kim's Contribution: Kim raised \$320 more than Alexandra, who raised \$430. \$430 + \$320 = \$750. Step 2: Maryam's Contribution: Maryam raised \$400 more than Sarah, who raised \$300. \$300 + \$400 = \$700. Step 3: Total Amount: To find the total, we add the amounts from all four girls: \$430 (Alexandra) + \$750 (Kim) + \$300 (Sarah) + \$700 (Maryam) = \$2180. Final Answer: \#\#\#\# 2180.''}\\[2pt]
\emph{Output:} \texttt{\{"style\_instruction": "Your answer should adopt this style: patient and encouraging, guiding the reader step by step like a supportive tutor."\}}\\[6pt]

\textbf{Your Turn}\\[-2pt]
\emph{Input Response:} ``\{response\}''\\
\emph{Output (Please directly return JSON only):}\\
\bottomrule
\end{tabularx}
\caption{Prompt for extracting Style constraints}
\label{tab:style_instruction_prompt}
\end{table*}

\begin{table*}[t]
\centering
\small
\setlength{\tabcolsep}{6pt}
\renewcommand{\arraystretch}{1.12}
\begin{tabularx}{\linewidth}{>{\raggedright\arraybackslash}X}
\toprule

You will be given a model response. Extract the core \emph{answer structure} it uses and rewrite it as a single, imperative instruction that can be attached to future prompts.\\[6pt]

\textbf{Instructions}\\[-2pt]
\textbullet\ Return \textbf{JSON only} with the key \texttt{"structure\_instruction"}.\\
\textbullet\ The value must be \textbf{one sentence} (\(\leq\) 25 words) starting with: \textbf{``Your answer should follow this structure: ''}.\\
\textbullet\ Summarize the \textbf{structural format} (e.g., step numbering, bullet points, table layout, paragraph style, equation blocks), not content or style.\\
\textbullet\ \textbf{No explanations, no extra keys, no text outside JSON.}\\
\textbullet\ If the format is implicit, infer the most plausible high-level structure.\\[6pt]

\textbf{Example}\\[-2pt]
\emph{Input Response:} \texttt{``Don't worry, you can think of it step by step. Step 1: Kim's Contribution: Kim raised \$320 more than Alexandra, who raised \$430. \$430 + \$320 = \$750. Step 2: Maryam's Contribution: Maryam raised \$400 more than Sarah, who raised \$300. \$300 + \$400 = \$700. Step 3: Total Amount: To find the total, we add the amounts from all four girls: \$430 (Alexandra) + \$750 (Kim) + \$300 (Sarah) + \$700 (Maryam) = \$2180. Final Answer: \#\#\#\# 2180.''}\\[2pt]
\emph{Output:} \texttt{\{"structure\_instruction": "Your answer should follow this structure: sequential steps explicitly labeled as 'Step \{\{number\}\}: \ldots'."\}}\\[6pt]

\textbf{Your Turn}\\[-2pt]
\emph{Input Response:} ``\{response\}''\\
\emph{Output (Please directly return JSON only):}\\
\bottomrule
\end{tabularx}
\caption{Prompt for extracting Structure constraints}
\label{tab:format_instruction_prompt}
\end{table*}

\begin{table*}[t]
\centering
\small
\setlength{\tabcolsep}{6pt}
\renewcommand{\arraystretch}{1.12}
\begin{tabularx}{\linewidth}{>{\raggedright\arraybackslash}X}
\toprule

\textbf{You are an editor.}\\[6pt]

\textbf{[Task]}\\[-2pt]
Expand the input paragraph to approximately \{TARGET\_WORDS\} words (±10\%) while preserving its original meaning exactly.\\[6pt]

\textbf{\#========================================\#}\\[4pt]

\textbf{[Input]}\\[-2pt]
type: \{TASK\_TYPE\} \quad \textit{\# one of: "math", "multi\_hop\_qa"}\\
paragraph: \{PARAGRAPH\}\\[6pt]

\textbf{\#========================================\#}\\[4pt]

\textbf{[Allowed expansion methods] (content-preserving only)}\\[-2pt]
\textbullet\ Rephrase with synonyms or equivalent constructions that restate the same constraints.\\
\textbullet\ Add discourse connectives and hedging/flow words that do not introduce requirements (e.g., ``in particular'', ``more specifically'', ``in other words'', ``please'', ``carefully'').\\
\textbullet\ Split or merge sentences; use parallel phrasing that reiterates existing points with equivalent wording.\\
\textbullet\ Expand abbreviations already present; add appositives that restate an existing noun phrase without adding attributes.\\
\textbullet\ Mild fluency padding (e.g., ``clearly'', ``precisely''), only if it does not add evaluation or new properties.\\[6pt]

\textbf{\#========================================\#}\\[4pt]

\textbf{[Global hard constraints] (must all hold)}\\[-2pt]
\textbf{1) No new information:}\\
\quad-- Do NOT add facts, entities, numbers, dates, locations, steps, options, claims, premises, constraints, caveats, or assumptions that are not already present.\\
\quad-- Do NOT change scope, difficulty, intent, or acceptance criteria.\\
\quad-- Do NOT add examples, citations, references, safety notes, or background knowledge.\\[4pt]
\textbf{2) Preserve every original detail AND the answer-format requirements:}\\
\quad-- Keep all entities, quantities, units, variables, symbols, equations/inequalities, and conditions unchanged in meaning.\\
\quad-- \textbf{Do not omit, weaken, or alter any explicit answer-format or output-style instructions present in the paragraph.}\\
\quad\ \ \ * Examples include (but are not limited to): ``Your final answer must begin with `\#\#\#\#' '', etc.\\
\quad\ \ \ * You may restate these format directives verbatim or with equivalent wording, but their requirements must remain exactly the same.\\
\quad\ \ \ * Do not introduce any new formatting requirements not in the input.\\
\quad-- Keep the same language as the input (English is expected).\\
\quad-- Maintain the same modality/tone (e.g., request/imperative/neutral).\\[4pt]
\textbf{3) Output format (your response to this prompt):}\\
\quad-- Return ONLY a JSON object with exactly these keys: \texttt{\{\{"expanded": string\}\}}\\
\quad-- No additional fields, preface, or commentary outside the JSON.\\[6pt]

\textbf{\#========================================\#}\\[4pt]

\textbf{[Type-specific constraints]}\\[-2pt]
\textbf{-- If type == ``math'':}\\
\quad* Do not alter the mathematical setup, variable names, symbols, equations, inequalities, or numeric values.\\
\quad* Do not change what is being asked (objective) or any given constraints, bounds, or domain assumptions.\\
\quad* You may add connective phrases or restate the same condition, but must not derive intermediate results or add hints.\\[2pt]
\textbf{-- If type == ``multi\_hop\_qa'':}\\
\quad* Preserve all entities, events, temporal and spatial qualifiers, and relations between them.\\
\quad* Do not add external facts, missing links, or bridging knowledge; only restate what is already present.\\
\quad* You may clarify with equivalent wording (e.g., restating a constraint in different words), but do not introduce new premises.\\[6pt]

\textbf{[Length target]}\\[-2pt]
Aim for \{TARGET\_WORDS\} words; acceptable range is ±10\%. If exact targeting would harm fidelity, prefer fidelity and stay inside the range.\\[6pt]

\textbf{\#========================================\#}\\[4pt]

Produce the JSON now.\\
\bottomrule
\end{tabularx}
\caption{Prompt for paraphrasing an instruction with a length constraint}
\label{tab:instruction_length}
\end{table*}

% \begin{table*}[t]
% \centering
% \small
% \setlength{\tabcolsep}{6pt}
% \renewcommand{\arraystretch}{1.12}
% \begin{tabular}{|p{0.18\linewidth}|p{0.76\linewidth}|}
% \hline
% \textbf{Variant} & \textbf{Instruction Structure (template)} \\ \hline
% A: Minimal concat &
% \texttt{\{constraint\}\\\{input\}} \\ \hline
% B: Solve-then-satisfy &
% \texttt{Your first priority is to produce the correct solution to the problem. After that, you should make every effort to meet the specified constraints.}\\
% & \texttt{Problem: \{input\}}\\
% & \texttt{Constraints: \{constraint\}} \\ \hline
% C: Scratch-then-format &
% \texttt{Step 1: Think through the complete solution on scratch paper.}\\
% & \texttt{Step 2: Based on your solution, write a final answer that fully meets the constraint.}\\
% & \texttt{Problem: \{input\}}\\
% & \texttt{Constraint: \{constraint\}} \\ \hline
% \end{tabular}
% \caption{Instruction-structure variants used in the ablation on prompt formatting.}
% \label{tab:instruction_structure}
% \end{table*}

\begin{table*}[t]
\centering
\small
\setlength{\tabcolsep}{6pt}
\renewcommand{\arraystretch}{1.12}
\begin{tabularx}{\linewidth}{>{\raggedright\arraybackslash}X}
\toprule

\textbf{[Instruction]}\\[-2pt]
Rani has ten more crabs than Monic, who has 4 fewer crabs than Bo. If Bo has 40 crabs, calculate the total number of crabs the three have together.\\
Your final answer must begin with `\#\#\#\#' and should appear as the last part of your response, in accordance with the required format. Use: \#\#\#\#\{Your Answer\}.\\[6pt]

\textbf{\#========================================\#}\\[4pt]
\textbf{[Constraints]}\\[-2pt]
\textbullet\ Ensure that the keyword `metal' is not present in your response.\\
\textbullet\ Apply this approach when formulating your answer: determine each person's quantity based on given relationships, then sum all quantities; show step-by-step calculations leading to the final total.\\
\textbullet\ Write at least 18 full sentences.\\
\textbullet\ Your answer should follow this style: structured and logical, using clear headings and step-by-step explanations.\\[6pt]

\textbf{\#========================================\#}\\[4pt]
\textbf{[Model Response]}\\[-2pt]
\texttt{\# Solving the Crab Problem: A Step-by-Step Analysis}\\
\texttt{\#\# Understanding the Problem}\\
Let me break down this problem systematically to find the total number of crabs owned by Rani, Monic, and Bo. We are given specific relationships between the quantities each person possesses, and we need to work through these relationships carefully.\\
\texttt{\#\# Given Information}\\
The problem provides us with the following key pieces of information. Bo has 40 crabs, which serves as our starting point. Monic has 4 fewer crabs than Bo, which means we need to subtract 4 from Bo's total. Rani has ten more crabs than Monic, so we'll need to add 10 to Monic's total.\\
\texttt{\#\# Step 1: Determining Bo's Crab Count}\\
Bo's crab count is explicitly stated in the problem. Bo has 40 crabs, which is our foundation for all subsequent calculations. This value will help us determine the quantities for both Monic and Rani.\\
\texttt{\#\# Step 2: Calculating Monic's Crab Count}\\
Now I need to find how many crabs Monic has. The problem states that Monic has 4 fewer crabs than Bo. Since Bo has 40 crabs, I subtract 4 from 40 to get Monic's total.\\
Monic's crabs = Bo's crabs $-$ 4\\
Monic's crabs = 40 $-$ 4\\
Monic's crabs = 36\\
Therefore, Monic has 36 crabs.\\
\texttt{\#\# Step 3: Calculating Rani's Crab Count}\\
Next, I need to determine Rani's crab count. The problem tells us that Rani has ten more crabs than Monic. Since we just calculated that Monic has 36 crabs, I add 10 to this amount.\\
Rani's crabs = Monic's crabs $+$ 10\\
Rani's crabs = 36 $+$ 10\\
Rani's crabs = 46\\
Therefore, Rani has 46 crabs.\\
\texttt{\#\# Step 4: Finding the Total}\\
Now that we have determined each person's individual crab count, we can calculate the total. We simply add together the crabs owned by Bo, Monic, and Rani.\\
Total crabs = Bo's crabs $+$ Monic's crabs $+$ Rani's crabs \\
\textcolor{red}{\textbf{Total crabs = 40 $+$ 36 $+$ 46 = 86}}\\
\texttt{\#\# Verification of Our Answer}\\
Let me verify this solution by checking our relationships. Bo has 40 crabs, Monic has 36 crabs (which is indeed 4 fewer than 40), and Rani has 46 crabs (which is indeed 10 more than 36). All relationships check out correctly, confirming our calculations are accurate.\\
The three individuals together have a combined total of 86 crabs.\\
\texttt{\#\#\#\#86}\\[6pt]

\textbf{\#========================================\#}\\[4pt]

\textbf{[Reference Answer]}\\[-2pt]
122\\

\bottomrule
\end{tabularx}
\caption{\textit{Reasoning Error} example. A failure case from Claude-Sonnet-4.5 on GSM8K with additional constraints. The erroneous step is highlighted in red.}
\label{tab:error_case}
\end{table*}

\begin{table*}[t]
\centering
\small
\setlength{\tabcolsep}{6pt}
\renewcommand{\arraystretch}{1.12}
\begin{tabularx}{\linewidth}{>{\raggedright\arraybackslash}X}
\toprule

\textbf{[Instruction]}\\[-2pt]
Write a C function \texttt{int are\_intervals\_intersecting(int a, int b, int c, int d)} to solve the following problem:\\
Determine if two closed intervals intersect.\\
Args: a, b represent the first closed interval $[a,b]$ where $0 \le a \le b \le 1000$;\\
\phantom{Args: }c, d represent the second closed interval $[c,d]$ where $0 \le c \le d \le 1000$.\\
Returns: 1 if the intervals intersect, 0 otherwise.\\[6pt]

\textbf{\#========================================\#}\\[4pt]
\textbf{[Constraints]}\\[-2pt]
\textbullet\ Do not let your code span more than 49 lines.\\
\textbullet\ Apply this approach when formulating your answer: apply logical conditions based on interval boundary comparisons to determine intersection.\\
\textbullet\ Please validate that your function or variable names include the keyword \texttt{``close''}.\\[6pt]

\textbf{\#========================================\#}\\[4pt]
\textbf{[Model Response]}\\[-2pt]
\texttt{/**}\\
\texttt{ * @brief Determines if two closed intervals intersect.}\\
\texttt{ *}\\
\texttt{ * This function applies logical conditions based on interval boundary comparisons}\\
\texttt{ * to determine if two closed intervals [a, b] and [c, d] have any common points.}\\
\texttt{ *}\\
\texttt{ * @param a The start point of the first closed interval [a, b].}\\
\texttt{ * @param b The end point of the first closed interval [a, b].}\\
\texttt{ * @param c The start point of the second closed interval [c, d].}\\
\texttt{ * @param d The end point of the second closed interval [c, d].}\\
\texttt{ * @return int 1 if the intervals intersect, 0 otherwise.}\\
\texttt{ */}\\
\texttt{int \textcolor{red}{\textbf{are\_close\_intervals\_intersecting}}(int a, int b, int c, int d) \{}\\
\texttt{\ \ \ \ // Two closed intervals [a, b] and [c, d] intersect if:}\\
\texttt{\ \ \ \ // 1. The end of the first interval (b) is not strictly before the start of the second (c). (b >= c)}\\
\texttt{\ \ \ \ // AND}\\
\texttt{\ \ \ \ // 2. The start of the first interval (a) is not strictly after the end of the second (d). (a <= d)}\\
\texttt{\ \ \ \ // If both conditions are true, the intervals intersect.}\\
\texttt{\ \ \ \ return (b >= c \&\& a <= d) ? 1 : 0;}\\
\texttt{\}}\\[6pt]

% \textbf{\#========================================\#}\\[4pt]

% \textbf{[Reference]}\\[-2pt]
% Correct output: the two intervals intersect if and only if $b \ge c$ and $a \le d$.\\

\bottomrule
\end{tabularx}
\caption{\textit{Output Specification Error} example. A failure case from Gemini-2.5-Flash on MCEval with additional constraints. The erroneous step is highlighted in red.}
\label{tab:code_case}
\end{table*}

\end{document}